%% file: main.tex
\documentclass{article}

\usepackage{times}
\usepackage{graphicx} % more modern
\usepackage{subfigure} 

\usepackage{natbib}

\usepackage{algorithm}
\usepackage{algorithmic}

\usepackage{hyperref}

\usepackage{tikz}			% tikz drawing package
\usepackage{enumitem}
\usepackage{listing}
\usepackage{float}
\usepackage{tabularx}
\usepackage{amsmath}
\usepackage{amssymb}
\usepackage[font={small}]{caption}
\usepackage{multirow}
\usepackage[bottom]{footmisc}

\newfloat{example}{thp}{}
\floatname{example}{Example}

\usetikzlibrary{arrows,decorations.pathmorphing,backgrounds,positioning,fit,matrix,calc,shapes.geometric,shapes.misc, positioning,decorations.pathreplacing}

\usepackage[accepted]{whi2017}

\icmltitlerunning{e-QRAQ: A Multi-turn Reasoning Dataset and Simulator with Explanations}
\begin{document} 

\twocolumn[
\icmltitle{e-QRAQ: A Multi-turn Reasoning Dataset and Simulator with Explanations}



% you can specify symbols, otherwise they are numbered in order

\begin{icmlauthorlist}
\icmlauthor{Clemens Rosenbaum}{ibm,umass}
\icmlauthor{Tian Gao}{ibm}
\icmlauthor{Tim Klinger}{ibm}

\end{icmlauthorlist}

\icmlaffiliation{ibm}{IBM T. J. Watson Research Center, Yorktown Heights, NY, USA}
\icmlaffiliation{umass}{University of Massachusetts Amherst, Amherst, MA, USA}

\icmlcorrespondingauthor{Clemens G. Rosenbaum}{cgbr@cs.umass.edu}

\icmlkeywords{Interpretable Machine Learning, Reasoning, Question Answering, Explanation, Dataset}

\vskip 0.3in
]



\printAffiliationsAndNotice{}  % leave blank if no need to mention equal contribution

\begin{abstract} 
In this paper we present a new dataset and user simulator e-QRAQ (explainable Query, Reason, and Answer Question) which tests an Agent's ability to read an ambiguous text; ask questions until it can answer a challenge question; and explain the reasoning behind its questions and answer.  The User simulator provides the Agent with a short, ambiguous story and a challenge question about the story. The story is ambiguous because some of the entities have been replaced by variables.  At each turn the Agent may ask for the value of a variable or try to answer the challenge question.  In response the User simulator provides a natural language explanation of why the Agent's query or answer was useful in narrowing  down the set of possible answers, or not. To demonstrate one potential application of the e-QRAQ dataset, %using the provided explanations as supervision, 
we train a new neural architecture based on End-to-End Memory Networks to successfully generate both predictions and partial explanations of its current understanding of the problem. We observe a strong correlation between the quality of the prediction and explanation. % Specifically we experiment with explanations which describe the current set of possible answers to the challenge question as well as which variables are relevant to query.

\end{abstract} 

\section{Introduction}
In recent years deep neural network models have been successfully applied in a variety of applications such as machine translation \cite{cho2014learning}, object recognition \cite{krizhevsky2012imagenet,he2016deep}, game playing \cite{mnih2015human}, dialog \cite{weston2016langlearning} and more.  However, their lack of interpretability makes them a less attractive choice when stakeholders must be able to understand and validate the inference process.  Examples include medical diagnosis, business decision-making and reasoning, legal and safety compliance, etc.  This opacity also  presents a challenge simply for debugging and improving model  performance.  For neural systems to move into realms where more transparent, symbolic models are currently employed, we must find mechanisms to ground neural computation in meaningful human concepts, inferences, and explanations.  One approach to this problem is to treat the explanation problem itself as a learning problem and train a network to explain the results of a neural computation.  This can be done either with a single network learning jointly to explain its own predictions or with separate networks for prediction and explanation.  Regardless, the availability of sufficient labelled training data is a key impediment.  In previous work \cite{guo2016learning} we developed a synthetic conversational reasoning dataset in which the User presents the Agent with a simple, ambiguous story and a challenge question about that story.  Ambiguities arise because some of the entities in the story have been replaced by variables, some of which may need to be known to answer the challenge question.  A successful Agent must reason about what the answers might be, given the ambiguity, and, if there is more than one possible answer, ask for the value of a relevant variable to reduce the possible answer set.  In this paper we present a new dataset e-QRAQ constructed by augmenting the QRAQ simulator with the ability to provide detailed explanations about whether the Agent's response was correct and why.  Using this dataset we perform some preliminary experiments, training an extended End-to-End Memory Network architecture \cite{sukhbaatar_end--end_2015} to jointly predict a response and a partial explanation of its reasoning.  We consider two types of partial explanation in these experiments: \emph{the set of relevant variables}, which the Agent must know to ask a relevant, reasoned question; and \emph{the set of possible answers}, which the Agent must know to answer correctly. We demonstrate a strong correlation between the qualities of the prediction and explanation. 

\section{Related Work}
Current interpretable machine learning algorithms for deep learning can be divided into two approaches: one approach aims to explain black box models in a model-agnostic fashion \cite{ribeiro2016model,turner2016model}; another  studies learning models, in particular deep neural networks, by visualizing for example the activations or gradients inside the networks \cite{zahavy2016graying,shrikumar2016not,selvaraju2016grad}. Other work has studied the interpretability of traditional machine learning algorithms, such as decision trees \cite{hara2016making}, graphical models \cite{kim2015mind}, and learned rule-based systems \cite{malioutov2013exact}. Notably, none of these algorithms produces natural language explanations, although the rule-based system is  close to a human-understandable form if the features are interpretable. We believe one of the major impediments to getting NL explanations is the lack of datasets containing supervised explanations.

Datasets have often accelerated the advance of machine learning in their perspective areas \cite{ferraro2015survey}, including computer vision \cite{lecun1998mnist,krizhevsky2009learning,ILSVRC15,lin2014microsoft,krishna2016visual}, natural language \cite{lowe2015ubuntu,hermann2015teaching,dodge2015evaluating}, reasoning \cite{weston2015towards,bowman2015large,guo2016learning}, etc. Recently,  natural language explanation was added to complement  existing visual datasets via crowd-sourcing labeling \cite{reed2016learning}.  However, we know of no question answering or reasoning datasets which offer NL explanations. Obviously labeling a large number of examples with explanations is a difficult and tedious task -- and not one which is easily delegated to an unskilled worker. To make progress until such a dataset is available or other techniques obviate its need, we follow the approach of existing work such as \cite{weston2015towards, weston2016langlearning}, and generate synthetic natural language explanations from a simulator.

\section{The QRAQ Dataset}
A QRAQ domain, as introduced in \cite{guo2016learning}, has two actors, the User and the Agent.  The User provides a short story set in a domain similar to the HomeWorld domain of \cite{weston2015towards, narasimhan2015language} given as an initial context followed by a sequence of events, in temporal order, and a challenge question.  The stories are semantically coherent but may contain hidden, sometimes ambiguous, entity references, which the Agent must potentially resolve to answer the question.
\begin{example}[h]
  \centering
  \small
  \begin{enumerate}[itemsep=-1.5pt]
    \item[C1.] Hannah and Emma are in the office. 
    \item[C2.] John is in the park.
    \item[C3.] Bob and George are in the square.
    \item[E1.] Hannah picks up the gift.
    \item[E2.] \$v goes from the office to the park.
    \item[E3.] \$w goes from the park to the bank.
    \item[E4.] \$x goes from the office to the square.
    \item[E5.] Emma goes from the square to the bank.
    \item[E6.] \$y goes from the square to the bank.
    \item[Q:] Where is the gift?
  \end{enumerate}
\caption{A QRAQ Problem}
  \label{ex:QRAQ}
\end{example}
To do so, the Agent can query the User for the value of variables which hide the identity of entities in the story. At each point in the interaction, the Agent must determine whether it knows the answer, and if so, provide it; otherwise it must determine a variable to query which will reduce the potential answer set (a ``relevant'' variable).\\
In example \ref{ex:QRAQ} the actors \$v, \$w, \$x and \$y are treated as variables whose value is unknown to the Agent.  In the first event, for example, \$v refers to either Hannah or Emma, but the Agent can't tell which. In a realistic text this entity obfuscation might occur due to spelling or transcription errors, unknown descriptive references such as ``Emma's sibling'', or indefinite pronouns such as ``somebody''. Several datasets with 100k problems each and of varying difficulty have been released to the research community and are available for download \cite{qraq2016}.

\section{Explainable QRAQ: e-QRAQ}
\subsection{The Dataset}
This paper's main contribution is an extension to the original QRAQ simulator that provides extensive explanations of the reasoning process required to solve a QRAQ problem. These explanations are created dynamically at runtime, in response to the Agent's actions. The following two examples illustrate these explanations, for several different scenarios:
\begin{example}[h!]
  \centering
  \small
    \begin{itemize}[itemsep=-1.5pt]
        \item[] \textit{The problem:}
        \item[C1.] Silvia is in the porch.
        \item[C2.] Charles is in the cellar.
        \item[C3.] Maria is in the porch.
        \item[E1.] Charles goes from the cellar to the attic.
        \item[E2.] Charles goes from the attic to the terrace.
        \item[E3.] \$V0 goes from the porch to the boudoir.
        \item[Q:] Where is Maria
        \item[] \textit{The interaction:}
        \item[U$^*$:] Possible Answers: Porch, Boudoir; Relevant Variables: \$V0
        \item[A:] Who is \$V0?
        \item[U:] \$V0 is Silvia.
        \item[U:] This query was helpful, since it allowed the following inference:
        \newline We now know that \$V0 is Silvia, and not Maria. Maria can therefore not be in the boudoir.
        \item[U$^*$:] Possible Answers: Porch; Relevant Variables: $\emptyset$
        \item[A:] Maria is in the porch.
        \item[U:] This answer is correct.
    \end{itemize}
\caption{A successfully solved e-QRAQ Problem (with explanations)}
\label{ex:iHOP1}
\end{example}
\begin{example}[h!]
  \centering
  \small
    \begin{itemize}[itemsep=-1.5pt]
        \item[] \textit{The problem:}
        \item[C1.] Paul is in the attic.
        \item[C2.] Maria is in the cellar.
        \item[C3.] Charles is in the attic.
        \item[E1.] Maria goes from the cellar to the terrace.
        \item[E2.] \$V4 goes from the attic to the porch.
        \item[E3.] Maria goes from the terrace to the boudoir
        \item[Q:] Where is Charles?
        \item[] \textit{The interaction:}
        \item[U$^*$:] Possible Answers: Attic, Porch; Relevant Variables: \$V4
        \item[A:] Who is \$V1?
        \item[U:] This query was not helpful, since \$V1 does not even occur in the problem.
        \item[U$^*$:] Possible Answers: Attic, Porch; Relevant Variables: \$V4
        \item[A:] Charles is in the porch.
        \item[U:] This was a guess, since Charles could still have been \$V4, and thereby in the Porch or in the Attic.\newline This guess was correct.
    \end{itemize}
\caption{An unsuccessfully solved e-QRAQ Problem (with explanations)}
\label{ex:iHOP2}
\end{example}
% CR: edited the following section to talk on types of responses (U vs U*) instead of tokens 
The context (C), events (E), and question (Q) parts of the problem are identical to those in a QRAQ problem.  In addition there is a trace of the interaction of a trained Agent (A) model with the User (U) simulator.  The simulator provides two kinds of explanations in response to the Agent's query or answer.  The first kind denoted ``U'' indicates whether the Agent's response is correct or not and why. The second kind of explanation, denoted ``U$^*$'' provides a full description of what can be inferred in the current state of the interaction.  In this case the relevant information is the set of possible answers at different points in the interaction (Porch, Boudoir / Porch for Example \ref{ex:iHOP1}) and the set of relevant variables (\$V0 / none for Example \ref{ex:iHOP1}). 

In Example \ref{ex:iHOP1}, illustrating a successful interaction, the Agent asks for the value of \$V0 and the User responds with the answer (Silvia) as well as an explanation indicating that it was correct (helpful) and why.  Specifically, in this instance it was helpful because it enabled an inference which reduced the possible answer set (and reduced the set of relevant variables). On the other hand, in Example \ref{ex:iHOP2}, we see an example of a bad query and corresponding critical explanation. 

In general, the e-QRAQ simulator offers the following explanations to the Agent:
\paragraph{Answers} When answering, the User will provide feedback depending on whether or not the Agent has enough information to answer; that is, on whether the set of possible answers contains only one answer. If the Agent has enough information, the User will only provide feedback on whether or not the answer was correct and on the correct answer if the answer was false. If the agent does not have enough information, and is hence guessing, the User will say so and list all still relevant variables and the resulting possible answers.
\paragraph{Queries} When querying, the User will provide several kinds of feedback, depending on how useful the query was. A query on a variable not even occurring in the problem will trigger an explanation that says that the variable is not in the problem. A query on an irrelevant variable will result in an explanation showing that the story's protagonist cannot be the entity hidden by that variable. Finally, a useful (i.e. relevant) query will result in feedback showing the inference that is possible by knowing that variable's reference. This set of inference can also serve as the detailed explanation to obtain the correct answer above. 

The e-QRAQ simulator will be available upon publication of this paper at the same location as QRAQ \cite{qraq2016} for researchers to test their interpretable learning algorithms.

\subsection{The ``interaction flow''}
\begin{figure}[b!]
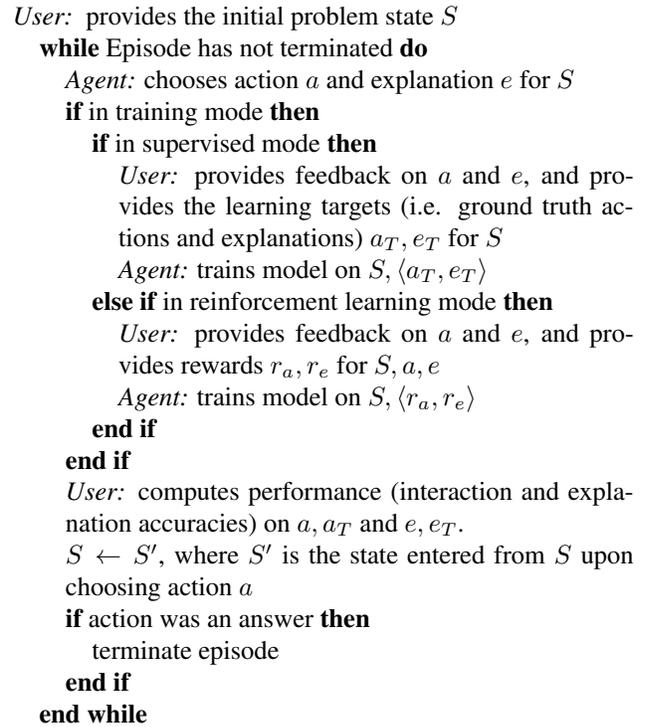

\begin{algorithmic}
\STATE[\textit{User:}] provides the initial problem state $S$
\WHILE{Episode has not terminated}
    \STATE \textit{Agent:} chooses action $a$ and explanation $e$ for $S$
    \IF{in training mode}
    \IF{in supervised mode}
        \STATE \textit{User:}  provides feedback on $a$ and $e$, and provides the learning targets (i.e. ground truth actions and explanations) $a_T, e_T$ for $S$
        \STATE \textit{Agent:} trains model on $S, \langle a_T, e_T \rangle$
    \ELSIF{in reinforcement learning mode}
        \STATE \textit{User:}  provides feedback on $a$ and $e$, and provides rewards $r_a, r_e$ for $S, a, e$
        \STATE \textit{Agent:} trains model on $S, \langle r_a, r_e \rangle$
    \ENDIF
    \ENDIF
    \STATE \textit{User:} computes performance (interaction and explanation accuracies) on $a, a_T$ and $e, e_T$.
    \STATE $S\leftarrow S'$, where $S'$ is the state entered from $S$ upon choosing action $a$
    \IF{action was an answer}
        \STATE terminate episode
    \ENDIF
\ENDWHILE
\end{algorithmic}
\caption{The User-Agent Interaction}
\label{fig:user-agent-interaction}
\end{figure}
The normal interaction flow between the User and the Agent during runtime of the simulator is shown in Figure \ref{fig:user-agent-interaction}, and is - with the exception of the additional explanations - identical to the interaction flow for the original QRAQ proglems \cite{guo2016learning}. This means that the User acts as a scripted counterpart to the Agent in the simulated e-QRAQ environment.
We show interaction flows for both supervised and reinforcement learning modes. Additionally, we want to point out that $e$ in Figure \ref{fig:user-agent-interaction} can be both U and U$^*$, i.e. both the natural language explanation and the internal state explanations.
Performance and accuracy are measured by the User, that compares the Agent's suggested actions and the Agent's suggested explanations with the ground truth known by the User.
\section{Experimental Setup}
\begin{figure}[h!]
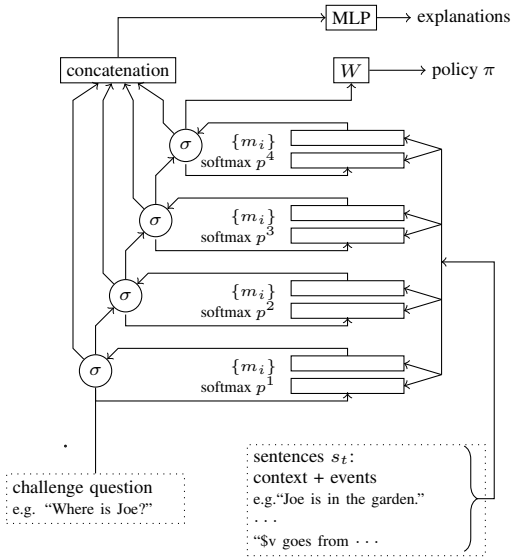

\include{architecture-drawing}
\caption{The modified E2E-Memory Network architecture simultaneously generating answers to the challenge question and explanations of its internal belief state, shown with four internal ``hops''.}
\label{architecture}
\end{figure}

For the experiments, we use the User simulator explanations to train an extended memory network. As shown in Figure \ref{architecture}, our network architecture extends the End-to-End Memory architecture of \cite{sukhbaatar_end--end_2015}, adding a two layer Multi-Layer Perceptron to a concatenation of all ``hops'' of the network. The explanation and response prediction are trained jointly.  In these preliminary experiments we do not train directly with the natural language explanation from U, just the explanation of what can be inferred in the current state U$^*$.  In future experiments we will work with the U explanations directly.
Specifically, for our experiments, we provide a classification label for the prediction output generating the Agent's actions, and a vector $x_e$ of the following form to the explanation output (where $oh_d(w)$ is an one-hot encoding of dimensionality (or vocabulary size) $d$ of word $w$, and $\mathcal{E}$ is the explanation set:
\begin{align}
    x_e     &= \sum_{w \in \mathcal{E}} oh_d(w)
    \intertext{We then train the network, using Adam \cite{kingma_adam:_2014}, on the combined loss (where $CE(y, \hat{y})$ is the cross-entropy between the true labels $y$ and the estimated labels $\hat{y}$, $\hat{x}_i$ is the network's interaction output and $\hat{x}_e$ is the networks explanation output):}
    \mathcal{L} &= CE(x_i, \hat{x}_i) + || x_e - \hat{x}_e||_2^2
\end{align}
\begin{figure}[b!]
\includegraphics[width=\columnwidth]{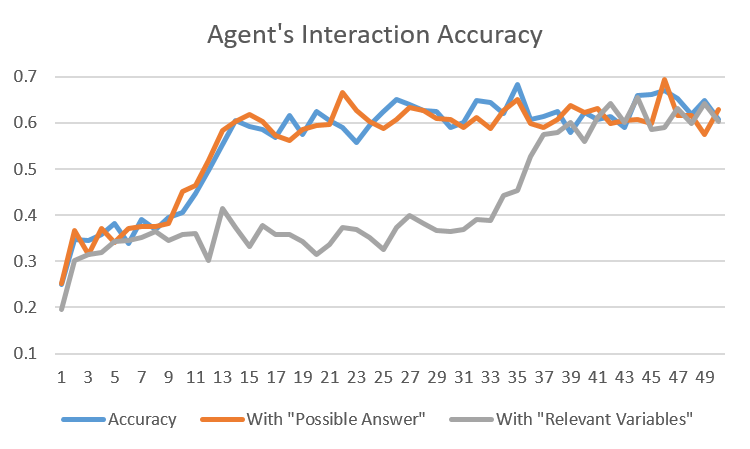}
\caption{The Interaction Accuracy (over 50 epochs with 1000 problems each)}
\label{fig:intacc}
\end{figure}

For testing, we consider the network to predict a entity in the explanation if the output vector $\hat{x}_e$ surpasses a threshold for the index corresponding to that entity. We tried several thresholds, some adaptive (such as the average of the output vector's values), but found that a fixed threshold of .5 works best.

\section{Results}

To evaluate the model's ability to jointly learn to predict and explain its predictions we performed two experiments. First, we investigate how the prediction accuracy is affected by jointly training the network to produce explanations. Second, we evaluate how well the model learns to generate explanations. To understand the role of the explanation content in the learning process we perform both of these experiments for each of the two types of explanation: relevant variables and possible answers. We do not perform hyperparameter optimization on the E2E Memory Network, since we are more interested in relative performance. While we only show a single experimental run in our Figures, results were nearly identical for over five experimental runs.

\begin{figure}[t!]
\includegraphics[width=\columnwidth]{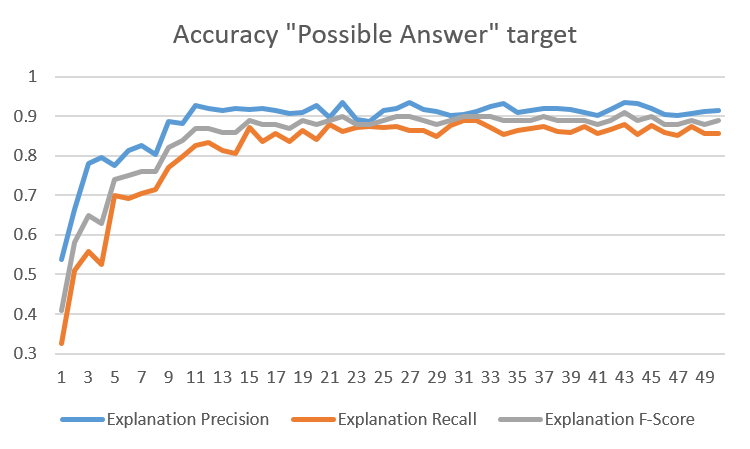}
\includegraphics[width=\columnwidth]{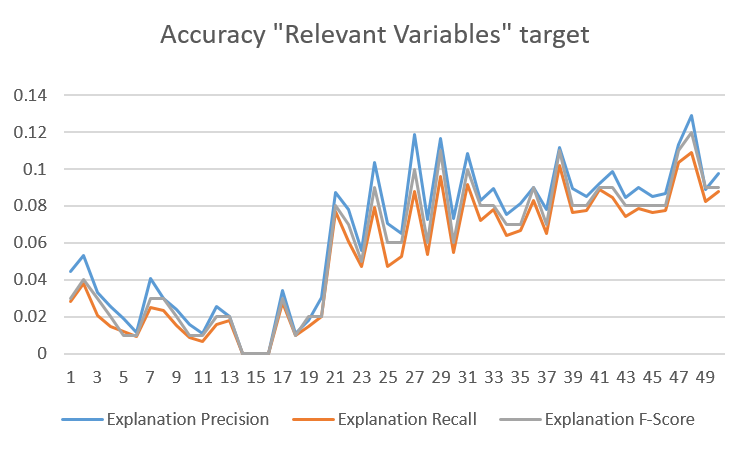}
\caption{The Explanation Accuracies (over 50 epochs with 1000 problems each)}
\label{fig:expacc}
\end{figure}
The experimental results differ widely for  the two kinds of explanation considered, where an explanation based on possible answers provides better scores for both experiments. As illustrated in Figure \ref{fig:intacc}, simultaneously learning possible-answer explanations does not affect prediction, while learning relevant-variable explanation learning severely impairs prediction performance, slowing the learning by roughly a factor of four. We can observe the same outcome for the quality of the explanations learned, shown in Figure \ref{fig:expacc}. Here again the performance on possible-answer explanations is significantly higher than for relevant-variable explanations. Possible-answer explanations reach an F-Score of .9, while relevant-variable explanations one of .09 only, with precision and recall only slightly deviating from the F-Score in all experiments.  

We would expect that explanation performance should correlate with prediction performance.  Since Possible-answer knowledge is primarily needed to decide if the net has enough information to answer the challenge question without guessing and relevant-variable knowledge is needed for the net to know what to query, we analyzed the network's performance on querying and answering separately. The memory network has particular difficulty learning to query relevant variables, reaching only about .5 accuracy when querying. At the same time, it learns to answer very well, reaching over .9 accuracy there. Since these two parts of the interaction are what we ask it to explain in the two modes, we find that the quality of the explanations strongly correlates with the quality of the algorithm executed by the network.

\section{Conclusion and Future Work}
We have constructed a new dataset and simulator, e-QRAQ, designed to test a network's ability to explain its predictions in a set of multi-turn, challenging reasoning problems. In addition to providing supervision on the correct response at each turn, the simulator provides two types of explanation to the Agent: A natural language assessment of the Agent's prediction which includes language about whether the prediction was correct or not, and a description of what can be inferred in the current state -- both about the possible answers and the relevant variables.  We used the relevant variable and possible answer explanations to jointly train a modified E2E memory network to both predict and explain it's predictions.  Our experiments show that the quality of the explanations strongly correlates with the quality of the predictions.  Moreover, when the network has trouble predicting, as it does with queries, requiring it to generate good explanations slows its learning. For future work, we would like to investigate whether we can train the net to generate natural language explanations and how this might affect prediction performance.

%\vfill\null
\bibliography{whi}
\bibliographystyle{icml2017}

\end{document}

%% file: architecture-drawing.tex
\begin{tikzpicture}
\scriptsize
		% THE MEMORY LOGIC
		\node[draw,dotted,rectangle,text width=30mm] (story) at (12, -.42) {sentences $s_t$:\newline context + events \newline\tiny  e.g.``Joe is in the garden.''\newline $\cdots$\newline ``\$v goes from $\cdots$};
		\node[draw,dotted,rectangle,text width=22mm] (question) at (8.4, -.4) {challenge question\newline \tiny  e.g. ``Where is Joe?''};
		\draw [decorate,decoration={brace,amplitude=5pt,mirror}] (13.3,-1.1) -- (13.3,.3);

		\draw (11, 1) rectangle (12.5, 1.2) {}; \node at (10.3, 1.1) {\tiny softmax $p^1$};
		\draw (11, 1.3) rectangle (12.5, 1.5) {}; \node at (10.5, 1.35) {\tiny $\{m_i\}$};
		
		\draw (11, 2) rectangle (12.5, 2.2) {}; \node at (10.3, 2.1) {\tiny softmax $p^2$};
		\draw (11, 2.3) rectangle (12.5, 2.5) {}; \node at (10.5, 2.35) {\tiny $\{m_i\}$};
				
		\draw (11, 3) rectangle (12.5, 3.2) {}; \node at (10.3, 3.1) {\tiny softmax $p^3$};
		\draw (11, 3.3) rectangle (12.5, 3.5) {}; \node at (10.5, 3.35) {\tiny $\{m_i\}$};
		
		\draw (11, 4) rectangle (12.5, 4.2) {}; \node at (10.3, 4.1) {\tiny softmax $p^4$};
		\draw (11, 4.3) rectangle (12.5, 4.5) {}; \node at (10.5, 4.35) {\tiny $\{m_i\}$};
		
		\draw[-] (13, 1.25) to (13,4.25);
		
		\draw[->] (13, 1.25) to (12.5,1.1);
		\draw[->] (13, 1.25) to (12.5,1.4);
		
		\draw[->] (13, 2.25) to (12.5,2.1);
		\draw[->] (13, 2.25) to (12.5,2.4);
		
		\draw[->] (13, 3.25) to (12.5,3.1);
		\draw[->] (13, 3.25) to (12.5,3.4);
		
		\draw[->] (13, 4.25) to (12.5,4.1);
		\draw[->] (13, 4.25) to (12.5,4.4);
		
		\draw[->] (13.47, -.4) |- (13.7, -.4) |- (13, 2.75);
		
		\node[draw,circle,scale=.1] (sigma0) at (8, 0.3) {$ $};
%		\node[draw,rectangle] (zi) at (8, 1) {$ $};
		
		\node[draw,circle] (sigma1) at (8.4, 1.3) {$\sigma$};
		\draw[->] (11.75, 1.5) |- ($(sigma1)+(.3,.3)$) to (sigma1);
				
		\node[draw,circle] (sigma2) at (8.8, 2.3) {$\sigma$};
		\draw[->] (11.75, 2.5) |- ($(sigma2)+(.3,.3)$) to (sigma2);
		
		\node[draw,circle] (sigma3) at (9.2, 3.3) {$\sigma$};
		\draw[->] (11.75, 3.5) |- ($(sigma3)+(.3,.3)$) to (sigma3);
		
		\node[draw,circle] (sigma4) at (9.6, 4.3) {$\sigma$};
		\draw[->] (11.75, 4.5) |- ($(sigma4)+(.3,.3)$) to (sigma4);
		
		\draw[->] (sigma1) |- ($(sigma2)+(-.4,-.4)$) to (sigma2);
		\draw[->] (sigma2) |- ($(sigma3)+(-.4,-.4)$) to (sigma3);
		\draw[->] (sigma3) |- ($(sigma4)+(-.4,-.4)$) to (sigma4);
		
		\draw[->] ($(sigma1)+(0,-.25)$) |- (11.75, .9) |- (11.75, 1);
		\draw[->] ($(sigma2)+(0,-.25)$) |- (11.75, 1.9) |- (11.75, 2);
		\draw[->] ($(sigma3)+(0,-.25)$) |- (11.75, 2.9) |- (11.75, 3);
		\draw[->] ($(sigma4)+(0,-.25)$) |- (11.75, 3.9) |- (11.75, 4);
%		\draw[->] (zi) to node[left,near end] {$z_i$} (8, 1.75);
		\node[draw,rectangle] (concat) at (8.7,5.3) {concatenation};
		
		\node[draw,rectangle] (w) at (11.8, 5.3) {$W$};
		
		\node (pol) at (13.3, 5.3) {policy $\pi$};
		\draw[->] (w) to (pol);
		
		\draw[-] (question) to (sigma1);
		
		\draw[->] (sigma1) to ($(sigma1)+(-.3, .3)$) to (8.1, 4.9) to (concat);
		\draw[->] (sigma2) to ($(sigma2)+(-.3, .3)$) to (8.5, 4.9) to (concat);
		\draw[->] (sigma3) to ($(sigma3)+(-.3, .3)$) to (8.9, 4.9) to (concat);
		\draw[->] (sigma4) to ($(sigma4)+(-.3, .3)$) to (9.3, 4.9) to (concat);
		\draw[->] (sigma4) to ($(sigma4)+(0., .6)$) |- ($(sigma4)+(2.2,.6)$) to (w);
		
		\node[draw,rectangle] (mlp) at (11.8, 6.) {MLP};
		\node (explanations) at (13.3, 6.) {explanations};
		\draw[->] (mlp) to (explanations);
		\draw[->] (concat) to ($(concat)+(0.,.7)$) to (mlp);
				
%		\draw[->] (7.9, 2.75) to (sum);
	\end{tikzpicture}